\documentclass[10pt,twocolumn,letterpaper]{article}

\usepackage{wacv}
\usepackage{times}
\usepackage{epsfig}
\usepackage{graphicx}
\usepackage{amsmath}
\usepackage{amssymb}
\usepackage{algorithm} 
\usepackage{algpseudocode}  

\usepackage{times,graphicx,amsmath,amssymb,booktabs,tabulary,multirow,overpic,xcolor,bbm,amssymb}
\usepackage[caption=false]{subfig}

\newcolumntype{x}[1]{>{\centering\arraybackslash}p{#1pt}}

\newlength\savewidth\newcommand\shline{\noalign{\global\savewidth\arrayrulewidth
  \global\arrayrulewidth 1pt}\hline\noalign{\global\arrayrulewidth\savewidth}}
\newcommand{\tablestyle}[2]{\setlength{\tabcolsep}{#1}\renewcommand{\arraystretch}{#2}\centering\footnotesize}
\makeatletter\renewcommand\paragraph{\@startsection{paragraph}{4}{\z@}
  {.5em \@plus1ex \@minus.2ex}{-.5em}{\normalfont\normalsize\bfseries}}\makeatother

\wacvfinalcopy 

\def\wacvPaperID{****} 

\ifwacvfinal\fi
\begin{document}

\title{Instance Segmentation with Point Supervision}

\author{
{Issam H. Laradji$^{1,2}$, Negar Rostamzadeh$^{1}$, Pedro O. Pinheiro$^{1}$, David Vazquez$^{1}$, Mark Schmidt$^{2,1}$}\\
{$^{1}$Element AI, Montreal, Canada} \hspace{1cm} $^2$University of British Columbia, Vancouver, Canada\\
{\hspace{-1.3cm}\{issamou,schmidtm\}@cs.ubc.ca} \hspace{1.5cm} {\{negar,pedro,dvazquez\}@elementai.com}
}

\maketitle
\ifwacvfinal\thispagestyle{empty}\fi

\begin{abstract}
Instance segmentation methods often require costly per-pixel labels. We propose a method that only requires point-level annotations. During training, the model only has access to a single pixel label per object, yet the task is to output full segmentation masks. To address this challenge, we construct a network with two branches: (1) a localization network (L-Net) that predicts the location of each object; and (2) an embedding network (E-Net) that learns an embedding space where pixels of the same object are close. The segmentation masks for the located objects are obtained by grouping pixels with similar embeddings. At training time, while L-Net only requires point-level annotations, E-Net uses pseudo-labels generated by a class-agnostic object proposal method. We evaluate our approach on PASCAL VOC, COCO, KITTI and CityScapes datasets. The experiments show that our method (1) obtains competitive results compared to fully-supervised methods in certain scenarios; (2) outperforms fully- and weakly- supervised methods with a fixed annotation budget; and (3) is a first strong baseline for instance segmentation with point-level supervision.
\end{abstract}

\section{Introduction}
\label{sec:intro}
Instance segmentation is the task of classifying every object pixel into a category and discriminating between individual object instances. It has a wide variety of applications such as autonomous driving~\cite{cordts2016cityscapes}, scene understanding~\cite{lin2014microsoft, everingham2010pascal}, and medical imaging~\cite{pohle2001segmentation}.

\begin{figure}
\label{fig:overview}
\begin{center}
\includegraphics[width=\linewidth]{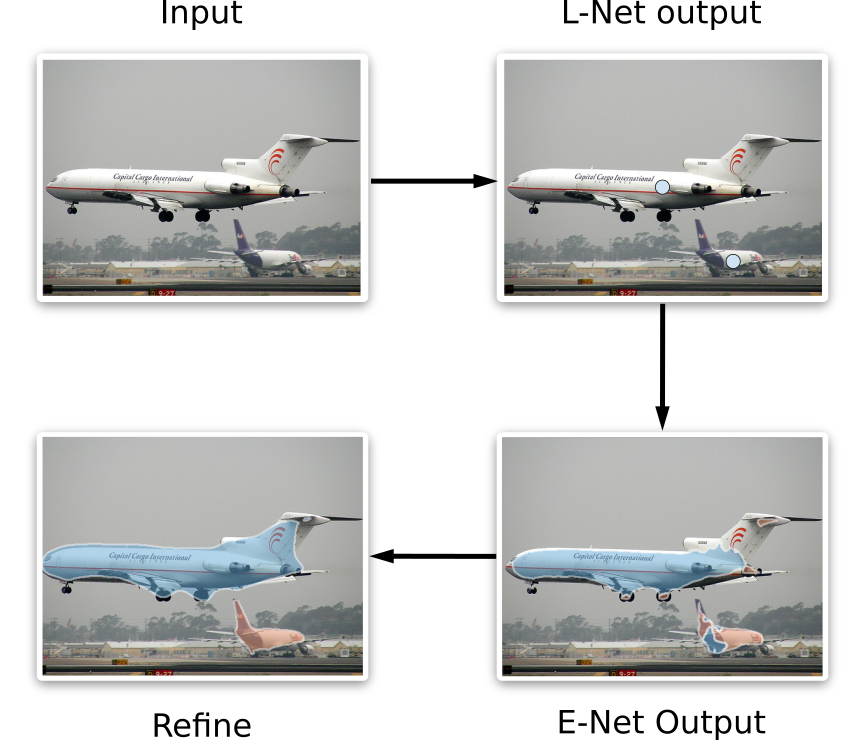}
\end{center}
\caption{{\bf WISE network.} Our method, WISE, is trained using point-level annotations only. At test time, WISE first uses L-Net to locate the objects in the image, and then uses E-Net to predict the masks of the located objects. Finally, the predicted masks are refined with the help of an object proposal method.}
\end{figure}

Most instance segmentation methods, such as Mask-RCNN~\cite{he2017mask} and MaskLab~\cite{chen2017masklab}, rely on per-pixel labels which requires huge human effort. For instance, obtaining labels for PASCAL VOC~\cite{everingham2010pascal} requires an average time of $239.7$ seconds per image~\cite{bearman2016s}. Other datasets with more objects to annotate such as CityScapes~\cite{cordts2016cityscapes} can take up to $1.5$ hours per image. 

\begin{figure}[t]
\begin{center}
\includegraphics[width=\linewidth]{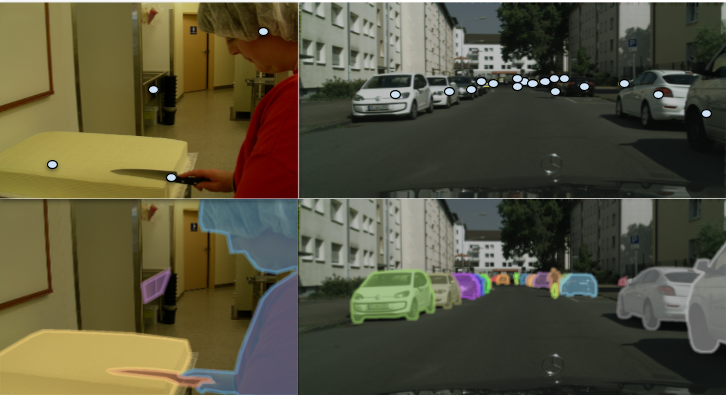}
\end{center}
\caption{\textbf{Image annotation.} Point-level (top) and per-pixel (bottom) labels for COCO and the CityScapes datasets.}
\label{fig:annotation}
\end{figure}

Indeed, having a method that can train with weaker supervision can vastly reduce the required annotation cost. According to Bearman~\emph{et al.}~\cite{bearman2016s}, manually collecting image-level and point-level labels for the PASCAL VOC dataset took only $20.0$ and $22.1$ seconds per image, respectively. These annotation methods are an order of magnitude faster than acquiring full segmentation labels (see Figure~\ref{fig:annotation} for a comparison between the point-level and per-pixel annotation methods).  

For semantic segmentation, other forms of weaker labels were explored such as bounding boxes~\cite{khoreva2017simple}, scribbles~\cite{lin2016scribblesup}, and image-level annotation~\cite{Zhou2018PRM}. For instance segmentation, few works exist that use weak supervision~\cite{Zhou2018PRM, cholakkal2019object}. In this paper, we propose a Weakly-supervised Instance SEgmentation (WISE) network, the first to address this task with point-level annotations.

WISE has two branches: (1) a localization network (L-Net) that predicts the location of each object; and (2) an embedding network (E-Net) that learns an embedding space where pixels of the same object are closer. L-Net is trained using a loss function that forces the network to output a single point per object instance. E-Net is trained using a similarity-based objective function to force the pixel embeddings to be similar within the same object mask. Since we do not have access to the ground-truth object masks, we instead use pseudo-masks generated by an object proposal method. These pseudo-masks belong to arbitrary objects and have no class labels and therefore cannot be directly applied for instance segmentation. At test time, L-Net first predicts the object locations. Second, E-Net outputs the embedding value for each pixel. Then the pixels with the most similar embeddings to an object's predicted pixel location become part of that object's mask (Figure~\ref{fig:overview}).

We summarize our contributions as follows: (1) we provide a first strong baseline for instance segmentation with point-level supervision; (2) we evaluate our method on a wide variety of datasets, including, PASCAL VOC~\cite{everingham2010pascal}, COCO~\cite{lin2014microsoft}, CityScapes~\cite{cordts2016cityscapes}, and KITTI~\cite{Geiger2012CVPR} datasets; (3) we obtain competitive results compared to fully-supervised methods; and (4) our method outperforms fully- and weakly- supervised methods when the annotation budget is limited.

\section{Related Work}
\label{sec:relatedwork}
Our approach lies at the intersection of object localization, metric learning, object proposal methods, and instance segmentation. These topics have been studied extensively and we review the literature below. The novelty of our method is the combination of these techniques into a new setup, namely, instance segmentation with point-level supervision.

\paragraph{Instance segmentation.}
Instance segmentation is an important computer vision task that can be applied in many real-life applications~\cite{ren2017end, romera2016recurrent}. This task consists of classifying every object pixel into categories and distinguishing between object instances. Most methods follow a two step procedure~\cite{he2017mask, chen2017masklab, fu2019retinamask}, where they first detect objects and then segment them. For instance, Mask-RCNN~\cite{he2017mask} uses Faster-RCNN~\cite{ren2015faster} for detection and an FCN network~\cite{long2015fully} for segmentation. However, these methods require dense labels which leads to a high annotation time for new applications.

\paragraph{Embedding-based instance segmentation.}
Another class of instance segmentation methods obtain the object masks by grouping pixels based on a similarity measure. Notable works in this category include methods based on watershed~\cite{bai2017deep}, template matching~\cite{uhrig2016pixel} and associative embedding~\cite{newell2017associative}. Fathi~\emph{et al.}~\cite{fathi2017semantic} propose a grouping-based method that first learns the object locations and then learns the pixel embeddings in order to distinguish between object instances. These methods also require per-pixel labels which are costly to acquire for new applications. However, our method follows a similar procedure for obtaining the segmentation masks while requiring weaker supervision.

\paragraph{Weakly supervised instance segmentation.}
Per-pixel labels used by fully supervised instance segmentation methods require high annotation cost~\cite{everingham2010pascal, cordts2016cityscapes}. Therefore many weakly supervised methods have been explored for object detection~\cite{Tang2018ECCV, Bilen2016CVPR}, semantic segmentation~\cite{pinheiro2015image, kolesnikov2016seed, ahn2018learning, selvaraju2017grad} and instance segmentation~\cite{khoreva2017simple, Zhou2018PRM, cholakkal2019object}. Point-level annotation is one of the fastest ways to annotate object instances, albeit one of the least informative forms of weak supervision. However, they were shown to be effective for semantic segmentation~\cite{bearman2016s}. Inspired by their cost-effectiveness, we explore the novel problem setup of instance segmentation with point-supervision in this work.

\paragraph{Object localization with point supervision.}
An important step in instance segmentation is to locate objects of interest before segmenting them. One way to perform object localization is to use object detection methods~\cite{ren2015faster, yolov3}. However, these methods require bounding-box labels. In contrast, several methods exist that use weaker supervision to identify object locations~\cite{song2014learning, song2014weakly, lempitsky2010learning, li2018csrnet}. Close to our work is LCFCN~\cite{laradji2018blobs} which uses point-level annotations in order to obtain the locations and counts of the objects of interest. While this method gives accurate counts and identifies a partial mask for each instance, it does not produce accurate segmentation of the instances. We extend this method by using an embedding network that groups pixels that are most similar to the predicted object locations in order to obtain their masks.

\begin{figure*}[t]
\begin{center}
\includegraphics[width=\linewidth]{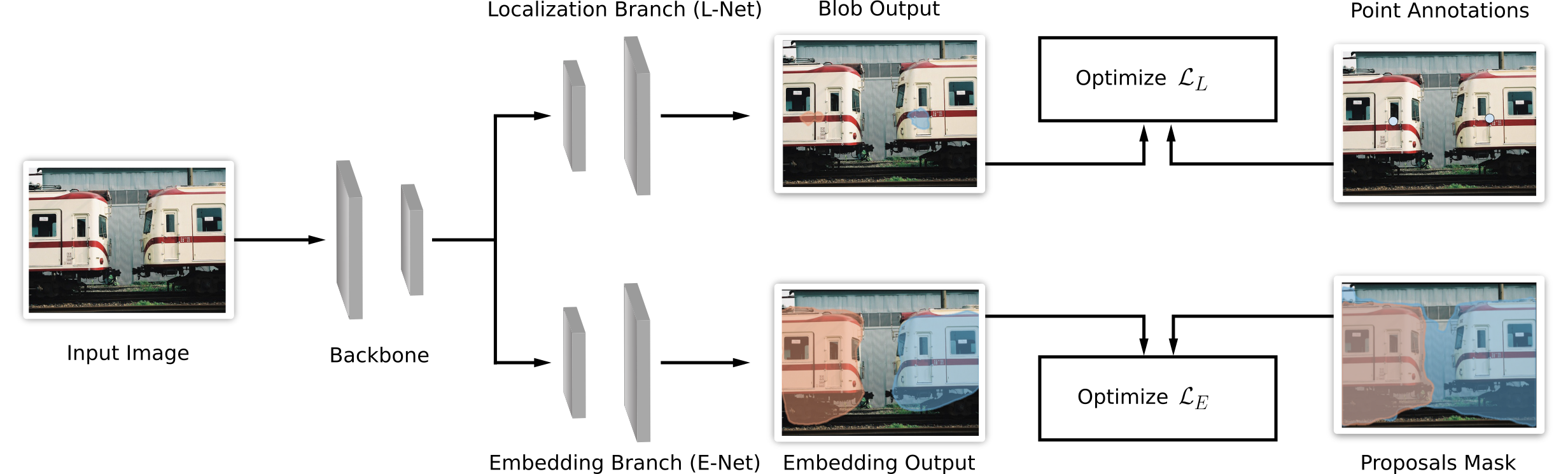}
\end{center}
\caption{{\bf Training WISE.} Our method consists of a localization branch (L-Net) and an embedding branch (E-Net). During training, L-Net optimizes Eq.~\ref{eq:loss_lcfcn} in order to output a single point per object instance. E-Net optimizes Eq.~\ref{eq:loss_enet} in order to group pixels that belong to the same object instance.}
\label{fig:WISE}
\end{figure*}

\paragraph{Object proposals.}
Weakly supervised methods often rely on object proposals~\cite{hosang2016makes} to ease the task of detection~\cite{Tang2018ECCV, Bilen2016CVPR}, and segmentation~\cite{pinheiro2015image, bearman2016s, Zhou2018PRM, kolesnikov2016seed}. Object proposals are class-agnostic methods that can output thousands of object candidates per image and have received great progress over the last decade~\cite{uijlings2013selective, zitnick2014edge, arbelaez2014multiscale, maninis2016convolutional, pinheiro2015learning, pinheiro2016learning}. SharpMask~\cite{pinheiro2016learning} is a popular deep-learning based object proposal method that has been successfully applied to many weakly supervised computer vision problems. However, their output object masks cannot be directly used for instance segmentation as they belong to arbitrary objects and have no class labels. Our framework uses pseudo-masks generated by SharpMask.

\section{Proposed Method}
\label{sec:method}

We address the problem of weakly-supervised instance segmentation, where each labeled object has a single point annotation. Our method, WISE network, has two output branches that share a common feature extraction backbone (Figure~\ref{fig:WISE}): (1) a localization branch (L-Net) that is trained for locating objects in the image, and (2) an embedding branch (E-Net) that outputs an embedding vector for each pixel. L-Net is trained using point-level annotations in order to output a single pixel for each object to represent its location and category in the image. On the other hand, E-Net is trained using pseudo-masks obtained by a pretrained proposal method. This allows E-Net to output an embedding vector for each pixel such that similar ones belong to the same object's pseudo-mask. Note that proposal methods have been widely used for different weakly-supervised problem setups~\cite{Zhou2018PRM, cholakkal2019object, Bilen2016CVPR, pinheiro2015image, bearman2016s}

WISE obtains the mask of an object as follows. First, L-Net outputs a pixel label per object to identify its location, category, and instance. Then, the embedding of every pixel in the image is compared to the embedding of the pixels predicted by L-Net to identify which object instance they belong to. Finally, the pixels are grouped to form the object masks in the image. 

\subsection{Localization Branch (L-Net)}
\label{ssec:lnet}
The goal of L-Net is to obtain the locations and categories of the objects in the image. L-Net is based on LC-FCN~\cite{laradji2018blobs} which trains with point level annotations to produce a single blob per object. While this was originally designed for counting, it is able to locate objects effectively. LC-FCN is based on a semantic segmentation architecture that is similar to FCN~\cite{long2015fully}. Indeed, semantic segmentation methods are not suitable for instance segmentation as they often predict large blobs that merge several object instances together. LC-FCN addresses this issue by optimizing a loss function that ensures that only a single small blob is predicted around the center of each object.

\begin{figure}[]
\begin{center}
\includegraphics[width=\linewidth]{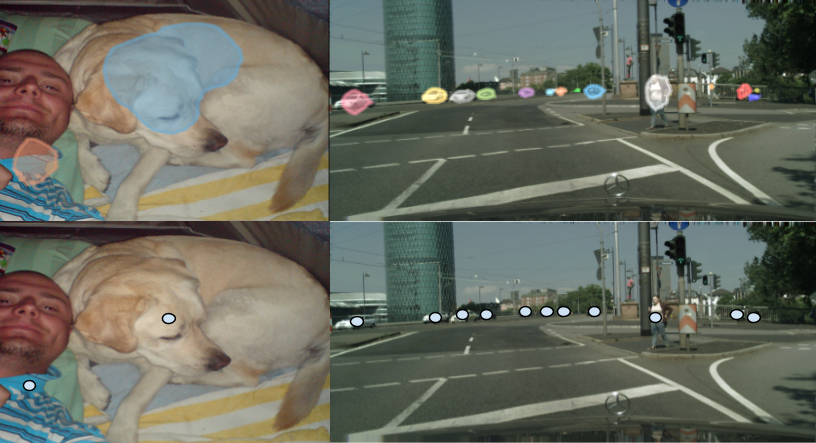}
\end{center}
\caption{{\bf Localization branch (L-Net).} L-Net's raw output is a small blob per predicted object (top). L-Net's final output is the set of pixels with the largest activation within their respective blobs (bottom). These pixels are used as input to E-Net at test time.}
\label{fig:localization}
\end{figure}

The location loss term $\mathcal{L}_L$ is described as:
\begin{equation}
\begin{split}
\mathcal{L}_L  &= \underbrace{\mathcal{L}_I(S,T)}_{\text{Image-level loss}} + \underbrace{\mathcal{L}_P(S,T)}_{\text{Point-level loss}}\\ &+ \underbrace{\mathcal{L}_S(S,T)}_{\text{Split-level loss}} + \underbrace{\mathcal{L}_F(S,T)}_{\text{False positive loss}}\;,
\label{eq:loss_lcfcn}
\end{split}
\end{equation}
where $T$ represent the point annotation ground-truth, and $S$ is LC-FCN's output mask. $\mathcal{L}_L$ consists of four terms: an image-level loss ($\mathcal{L}_I$) that trains the model to predict whether there is an object in the image; a point-level loss ($\mathcal{L}_P$) that encourages the model to predict a pixel for each object instance; a split-level ($\mathcal{L}_S$) and a false-positive ($\mathcal{L}_F$) loss that enforce the model to predict a single blob per instance (see~\cite{laradji2018blobs} for details for each of the loss components).
Since LC-FCN's predicted blobs are too small to be considered as useful segmentation masks, we instead leverage the location of each blob by identifying the pixel with the highest probability of being foreground (Figure~\ref{fig:localization}). 

\subsection{Embedding Branch (E-Net)}
\label{sec:enet}
The goal of E-Net is to produce object masks by grouping pixels with similar embeddings together. E-Net's architecture is based on FCN8~\cite{long2015fully}, which can output an embedding vector per image pixel. Using a similarity loss, E-Net learns to output similar embeddings for pixels that belong to the same object and dissimilar otherwise. This loss requires several points per object (including the background) in order to distinguish between different objects. While we do not have access to the ground-truth masks, we instead use pseudo-masks generated by an object proposal method to assign a mask for each object.

E-Net learns a mapping from an input image to a set of embedding vectors of size $d$ for each pixel. Let $E_i$ and $E_j$ be the embeddings for pixel $i$ and pixel $j$, respectively. We measure the similarity between a pair of pixels using a squared  exponential kernel function, similar to that of Fathi \emph{et al.}~\cite{fathi2017semantic}:
\begin{equation}
S(i, j) = \exp\Bigg({-\frac{||E_i-E_j||_2^2}{2d}}\Bigg)\;,
\label{eq:similarity}
\end{equation}
where $S(E_i, E_j)$ tends to $1$ as $E_i$ and $E_j$ get closer, and tends to $0$ as they get farther in the embedding space. Note that our method can use other similarity functions as in~\cite{newell2017associative, fathi2017semantic, kong2018grouppixels}.

Our goal is to train E-Net such that embeddings of pixel pairs belonging to the same object instance (\ie $y_i = y_j$) have the same embedding (\ie $S(i, j) = 1$) and to different object instances (\ie $y_i \neq y_j$) have different embeddings (\ie $S(i, j) = 0$). Therefore, E-Net minimizes the following loss function\footnote{Note that the $\log$ and $\exp$ cancel out in the first term of the equation but not the second term.}: 
\begin{align}
\begin{split}
\mathcal{L}_E = -\sum_{(i,j) \in P} \Big[&\mathbbm{1}_{\{y_i=y_j\}} \log{S(E_i, E_j)}  \;\;+ \\&
 \mathbbm{1}_{\{y_i\neq y_j\}} \log{(1  - S(E_i, E_j))}\Big]\;,
\label{eq:loss_enet}
\end{split}
\end{align}
where $P$ is a set of pixel pairs.

\begin{figure}[t]
\begin{center}
\includegraphics[width=\linewidth]{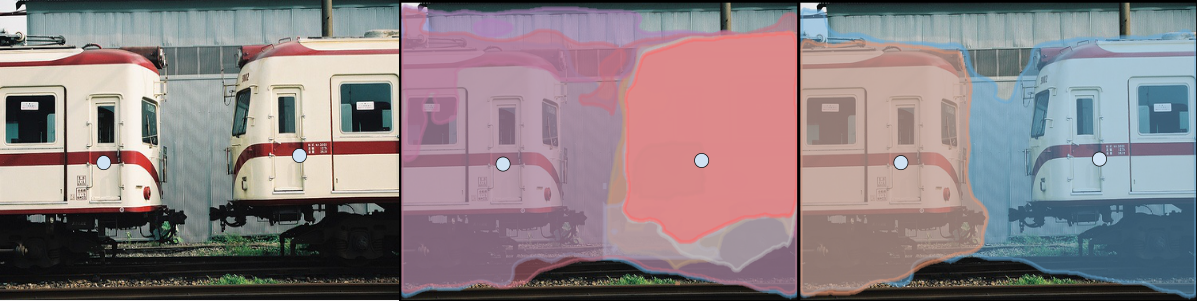}
\end{center}
\caption{\textbf{Pseudo-mask labels}. (Left) ground-truth point-level annotations; (Center) a set of generated object proposals that intersect with the point annotations; (Right) proposals with best ``objectness".}
\label{fig:gt}
\end{figure}

Since we require more than one point label per object to optimize Equation~\ref{eq:loss_enet}, we use extra points from pseudo-masks generated by an object proposal method (see Figure~\ref{fig:gt}). At each training iteration, the pseudo-mask of an object is randomly selected from the set of proposals (obtained by the proposal method) that intersect with the object's point annotation.  Further, we define the background as the region that does not contain any proposal mask. 

We obtain the set of pixel pairs $P$ for Eq.~\ref{eq:loss_enet} as follows. We pair each pixel represented by the point-level annotation with $k$ random pixels\footnote{We chose $k$ as the number of objects in the image.} from each object's pseudo-mask including the background region. This randomness allows the model to learn the important pixels that correspond to the objects of interest. The final objective function of WISE is defined as:
\begin{equation}
\mathcal{L}_W = \lambda\cdot \mathcal{L}_L + (1-\lambda)\cdot \mathcal{L}_E\;,
\label{eq:loss_wisenet}
\end{equation}
where $\lambda$ is the weight that balances between L-Net's and E-Net's loss terms.

\subsection{Prediction at Test Time}
\label{sec:prediction}
WISE predicts masks of objects using the following steps. First, L-Net outputs a pixel coordinate for each object representing its location and category. Second, E-Net outputs the embedding vectors for all pixels in the image. Third, we compute the similarity (Equation~\ref{eq:similarity}) between each pixel in the image and two sets of pixels: (1) L-Net's predicted pixel coordinates, and (2) several selected background pixels. Next, we assign each pixel to the most similar object, resulting in a mask for each object including the background region. Finally, the object masks are refined by replacing them with the pseudo-mask (generated from a proposal method) with the largest Jaccard similarity (see Figure~\ref{fig:overview}). 

For selecting the background pixels deterministically, we first define the background regions as the pixels that do not correspond to any of the generated proposal masks. We use the $k$-means algorithm for clustering the pixels embeddings into $k$ groups. Then, for each cluster we select the closest pixel to the mean of that cluster, giving us $k$ representative pixels from the background.

\section{Experiments}
\label{sec:experiments}
We evaluate the WISE network on a wide variety of datasets: PASCAL VOC~\cite{everingham2010pascal}, COCO~\cite{lin2014microsoft}, CityScapes~\cite{cordts2016cityscapes}, and KITTI~\cite{Geiger2012CVPR} datasets. We compare our results against fully-supervised, and weakly-supervised methods. We compare WISE against several baselines to showcase the efficacy of each of its components. We also fix the annotation budget for acquiring per-pixel, point-level, and image-level labels and compare several models based on the type of label they require. Unless otherwise specified, the performance is measured using average precision (AP) as in~\cite{he2016deep}, computed with Intersection-over-Union (IoU) thresholds of 0.25, 0.5, and 0.75. 

\subsection{Methods and Baselines}
We include the following methods in our benchmarks:

\noindent {\bf L-Net + Blobs:} use the raw output of L-Net (see Figure~\ref{fig:localization}) (which is a predicted blob per object in the scene) as mask prediction.

\noindent {\bf L-Net + Best proposal:} replace each object location predicted by L-Net with the SharpMask's proposal that has the highest ``objectness'' score.

\noindent {\bf L-Net + Oracle proposal:} replace each object location predicted by L-Net with the SharpMask's proposal that achieves the highest evaluation score (\eg mAP).

\noindent {\bf L-Net + GT-Mask:}
replace each object location predicted by L-Net with the ground-truth mask.

\noindent {\bf PRM + E-Net:}
use the object locations predicted by PRM (as described in~\cite{Zhou2018PRM}) as input to E-Net to obtain the object masks. Note that PRM only requires image-level labels.

\noindent {\bf GT-points + E-Net:} use the ground-truth object locations (point-level annotations) as input to E-Net to obtain the object masks. 

\noindent {\bf WISE (L-Net + E-Net):} use L-Net's predicted object locations as input to E-Net to obtain the object masks. 

\begin{table}[t]
    \centering
    \begin{tabular}{l|x{22}x{22}x{22}}
        \textbf{Method} &\textbf{AP$_{25}$} &\textbf{AP$_{50}$} &\textbf{AP$_{75}$}\\
        \shline
        L-Net + Blobs            & 08.4 & 01.2 & 00.1\\ 
        L-Net + Best proposal    & 42.9 & 33.4 & 19.1\\ 
        L-Net + Oracle proposal  & 57.3 & 45.1 & 37.2\\ 
        L-Net + GT-Mask          & 61.2 & 61.2 & 61.2\\ 
        \hline
        PRM + E-Net              & 43.0 & 32.0 & 19.0\\ 
        GT-points + E-Net        & 63.1 & 47.0 & 26.3\\ 
        \hline
        WISE (L-Net + E-Net) & 53.5 & 43.0 & 25.9\\ 
    \end{tabular}
    \vspace{3mm}
    \caption{\textbf{Ablation Studies}. A benchmark illustrating the contribution of each WISE's component on PASCAL VOC 2012.}
    \label{tab:pascal_baselines}
\end{table}

\subsection{Implementation Details}
L-Net and E-Net share the same backbone, a ResNet-50~\cite{he2016deep} pretrained on ImageNet~\cite{deng2009imagenet}. They also have independent upsampling paths with similar architecture as FCN8~\cite{long2015fully}. The number of output channels for L-Net is the number of classes, and for E-Net is $d=64$, the size of a pixel's embedding vector. We observed minor differences in the results between different embedding dimensions. For each image, we use $1000$ pretrained SharpMask~\cite{pinheiro2016learning} proposals (note that we do not finetune the proposal on any dataset). During training, for each point-annotation we sample a proposal non-uniformly based on its ``objectness'' score to represent its pseudo-mask.
We set $k$ as the number of predicted objects (by L-Net) for selecting the background pixels at test time. The model is trained using Adam~\cite{kingma2014adam} optimizer with a learning rate of $10^{-5}$ and a weight decay of $0.0005$ for $200k$ iterations with a batch size of $1$. We choose $\lambda=0.1$ in Equation~\ref{eq:loss_wisenet} in order to make the scale between its two loss terms similar.

\begin{table}[t]
    \centering
    \resizebox{\linewidth}{!}{%
    \begin{tabular}{l|x{47}x{16}x{16}x{16}}
        \textbf{Method}  & \textbf{Annotation} &\textbf{AP$_{25}$} &\textbf{AP$_{50}$} &\textbf{AP$_{75}$}\\
        \shline
        Mask R-CNN~\cite{zhu2017soft} & per-pixel    & 17.1 & 11.2 & 03.4\\
        SPN~\cite{zhu2017soft}        & image-level  & 26.0 & 13.0 & 04.0\\
        PRM~\cite{Zhou2018PRM}        & image-level  & 44.0 & 27.0 & 09.0\\
        Cholakkal~\emph{et al.}~\cite{cholakkal2019object}& image-level   & \textbf{48.5} & 30.2 & 14.4\\
        \hline
        PRM + E-Net (Ours)            & image-level  & 43.0 & 32.0 & 19.0\\
        WISE (Ours)                   & point-level  & 47.5 & \textbf{38.1} & \textbf{23.5}\\
    \end{tabular}
    }
    \vspace{3mm}
    \caption{\textbf{PASCAL VOC 2012 with a fixed annotation budget.} Comparison across methods with the same annotation budget.}
    \label{tab:pascal_budget}
\end{table}

\subsection{Experiments on PASCAL VOC 2012}
PASCAL VOC 2012~\cite{everingham2010pascal} contains $1,464$ and $1,449$ images for training and validation respectively, where objects come from $20$ categories. We use the point-level annotations provided by Bearman~\emph{et al.}~\cite{bearman2016s} as ground-truth for training our methods. We report the AP across several thresholds on the validation set, as described in the dataset's instance segmentation setup~\cite{everingham2010pascal}.

\begin{table*}[t]
    \centering
    \tablestyle{4pt}{1.05}
    \resizebox{\textwidth}{!}{%
    \begin{tabular}{l|x{12}x{12}x{12}x{12}x{12}x{12}x{12}x{12}x{12}x{12}x{12}x{12}x{12}x{12}x{12}x{12}x{12}x{12}x{12}x{12}|x{12}x{12}}
        \multicolumn{1}{l|}{\textbf{Method}}  &
        \multicolumn{1}{l|}{\textbf{\rotatebox[origin=c]{90}{plane}}} &
        \multicolumn{1}{l|}{\textbf{\rotatebox[origin=c]{90}{bike}}} &
        \multicolumn{1}{l|}{\textbf{\rotatebox[origin=c]{90}{bird}}} &
        \multicolumn{1}{l|}{\textbf{\rotatebox[origin=c]{90}{boat}}} &
        \multicolumn{1}{l|}{\textbf{\rotatebox[origin=c]{90}{bottle}}} &
        \multicolumn{1}{l|}{\textbf{\rotatebox[origin=c]{90}{bus}}} &
        \multicolumn{1}{l|}{\textbf{\rotatebox[origin=c]{90}{car}}} &
        \multicolumn{1}{l|}{\textbf{\rotatebox[origin=c]{90}{cat}}} &
        \multicolumn{1}{l|}{\textbf{\rotatebox[origin=c]{90}{chair}}} &
        \multicolumn{1}{l|}{\textbf{\rotatebox[origin=c]{90}{cow}}} &
        \multicolumn{1}{l|}{\textbf{\rotatebox[origin=c]{90}{table}}} &
        \multicolumn{1}{l|}{\textbf{\rotatebox[origin=c]{90}{dog}}} &
        \multicolumn{1}{l|}{\textbf{\rotatebox[origin=c]{90}{horse}}} &
        \multicolumn{1}{l|}{\textbf{\rotatebox[origin=c]{90}{motor}}} &
        \multicolumn{1}{l|}{\textbf{\rotatebox[origin=c]{90}{person}}} &
        \multicolumn{1}{l|}{\textbf{\rotatebox[origin=c]{90}{plant}}} &
        \multicolumn{1}{l|}{\textbf{\rotatebox[origin=c]{90}{sheep}}} &
        \multicolumn{1}{l|}{\textbf{\rotatebox[origin=c]{90}{sofa}}} &
        \multicolumn{1}{l|}{\textbf{\rotatebox[origin=c]{90}{train}}} &
        \textbf{\rotatebox[origin=c]{90}{tv}} &
        \textbf{\rotatebox[origin=c]{90}{Avg.}}  \\\shline
        \scriptsize SDS~\cite{hariharan2014simultaneous}        & 58.8 & 0.5 & 60.1 & 34.4 & 29.5 & 60.6 & 40.0 & 73.6 & 6.5 & 52.4 & 31.7 & 62.0 & 49.1 & 45.6 & 47.9 & 22.6 & 43.5 & 26.9 & 66.2 & 66.1 & 43.8  \\
        \scriptsize Chen \emph{et al.}~\cite{chen2015multi}     & 63.6 & 0.3 & 61.5 & 43.9 & 33.8 & 67.3 & 46.9 & 74.4 & 8.6 & 52.3 & 31.3 & 63.5 & 48.8 & 47.9 & 48.3 & 26.3 & 40.1 & 33.5 & 66.7 & 67.8 & 46.3  \\ 
        \scriptsize PFN~\cite{liang2015proposal}                & 76.4 & 15.6& 74.2 & 54.1 & 26.3 & 73.8 & 31.4 & 92.1 & 17.4& 73.7 & 48.1 & 82.2 & 81.7 & 72.0 & 48.4 & 23.7 & 57.7 & 64.4 & 88.9 & 72.3 & 58.7  \\
        \scriptsize R2-IOS~\cite{liang2016reversible}           & 87.0 & 6.1 & 90.3 & 67.9 & 48.4 & 86.2 & 68.3 & 90.3 & 24.5 & 84.2 & 29.6 & 91.0 & 71.2 & 79.9& 60.4 & 42.4 & 67.4 & 61.7 & 94.3 & 82.1 & 66.7  \\ 
        \scriptsize Fathi \emph{et al.}~\cite{fathi2017semantic}& 69.7 & 1.2 & 78.2 & 53.8 & 42.2 & 80.1 & 57.4 & 88.8 & 16.0 & 73.2 & 57.9 & 88.4 & 78.9 & 80.0 & 68.0 & 28.0 & 61.5 & 61.3 & 87.5 & 70.4 & 62.1  \\ \hline
        \scriptsize WISE (Ours)                             & 59.0 & 5.6 & 63.6 & 41.4 &21.9 & 40.6 & 34.1 & 73.8 & 8.5 & 38.7 & 29.1& 64.6 &58.1 & 60.4 & 33.3 & 25.1 &43.8 & 32.7 & 64.7 & 60.7 & 43.0   \\ 
    \end{tabular}}
    \vspace{3mm}
    \caption{\textbf{Comparison to fully supervised methods.} Per-class comparison against the AP$_{50}$ metric on PASCAL VOC 2012.}
    \label{tab:pascal_full}
\end{table*}

\begin{table*}[t]
    \centering
    \tablestyle{4pt}{1.05}
    \begin{tabular}{l|x{22}x{22}x{22}|x{22}x{22}x{22}|x{22}x{22}x{22}}
        \multirow{2}{*}{{\textbf{Model}}} & 
        \multicolumn{3}{c|}{\textbf{COCO 2014}} &\multicolumn{3}{c|}{\textbf{KITTI}} & \multicolumn{3}{c}{\textbf{CityScapes}}\\
        &\textbf{AP$_{25}$} & \textbf{AP$_{50}$}&\textbf{AP$_{75}$} &\textbf{AP$_{25}$} &\textbf{AP$_{50}$} &\textbf{AP$_{75}$} &\textbf{AP$_{25}$} &\textbf{AP$_{50}$} &\textbf{AP$_{75}$} \\ \shline
        L-Net Best proposal & 18.3 & 13.6 & 7.3 & 46.4 & 38.1 & 22.2 & 27.2 & 15.5 & 6.7\\
        WISE (Ours)     & 25.8 & 17.6 & 7.8 & 63.4 & 49.8 & 30.9 & 28.7 & 18.2 & 8.8\\
    \end{tabular}
    \vspace{3mm}
    \caption{\textbf{Baseline comparisons.} Results across different average precision IoU thresholds.}
    \label{tab:other_baseline}
\end{table*}

\subsubsection{Comparison to methods and baselines.}\label{sec:baselines}
In this section, we discuss the results shown in Table~\ref{tab:pascal_baselines}. A straightforward method to obtain object masks is to use L-Net's raw output (which we refer to as ``L-Net + Blobs''). However, it performs poorly as the predicted blobs are often small around the center of the object. 

A natural extension is to replace L-Net's predicted blobs by a segment proposal obtained from an object proposal method. 
Therefore, we discovered a reasonable strategy which is to replace each of L-Net's predicted blobs by the proposal of highest ``objectness'' score (``L-Net + Best-proposal''). However, ``L-Net + Oracle'' shows that a perfect proposal selection strategy can vastly improve on the segmentation results. 

Accordingly, we propose WISE which improves on ``L-Net + Best-proposal'' by having E-Net that learns rough segmentation of the objects. This allows to select better proposals by choosing those with the highest IoU. Note that other object proposal selection strategies have been used in other weakly supervised instance segmentation setups~\cite{Zhou2018PRM, cholakkal2019object}. 

To assess how much improvement we can make over L-Net, we report the results of ``GT-points + E-Net'' which uses the ground-truth points instead of L-Net's predictions. We see that L-Net's performance is close to its upper-bound. Further, we provide the results of ``PRM + E-Net'' which is an extension to WISE that can train using image-level annotations only. Similarly, we observe that the results are not widely different. However, image-level labels might not be suitable for datasets when the number of objects in an image is dense and when the same object class exist in almost every image as the \emph{car} category in CityScapes.

\subsubsection{Comparison to Similar Annotation Time}\label{sec:budget}
We compare the performance between state-of-the-art methods in Table~\ref{tab:pascal_budget} when the annotation time is fixed. Therefore, we limit the annotation budget to around $8.13$ hours which is calculated as $20.0 \times 1,464$ seconds. Bearman~\emph{et al.}~\cite{bearman2016s} has shown that it takes $20.0$, $22.1$, and $239.7$ seconds per image for collecting image-level, point-level, and per-pixel labels, respectively. As a result, for the same annotation time budget, we acquire $1,464$ images with image-level labels, $1,325$ images with point-level labels, and $122$ images with per-pixel labels. We selected these images uniformly without replacement from the training set. We also reported the result of  Mask R-CNN~\cite{massa2018mrcnn} trained on the images with the per-pixel labels. The table shows that our method significantly outperforms other approaches, suggesting that using point-level annotations is a cost-effective labeling method for instance segmentation. Further, Figure~\ref{fig:qualitative} illustrates that WISE can capture high quality masks for PASCAL VOC objects, although it can fail in merging two masks of the same object such as in the horse image.

\subsubsection{Comparison to Weakly and Fully Supervised Methods}
\label{sec:full}
Acquiring point-level labels is almost as cheap as image-level labels, yet they vastly improve results, as shown in Table~\ref{tab:pascal_budget}. For a fair evaluation, we compare ``PRM + E-Net'' which uses image-level labels against current state-of-the-art image-level instance segmentation methods. The concurrent work of~\cite{cholakkal2019object} performs better with respect to AP$_{25}$, which is expected as their counting results is better than LCFCN which is what L-Net is based on.
 
Further, we report WISE results against fully supervised methods in Table~\ref{tab:pascal_full} for each category with respect to AP$_{50}$. While WISE achieves competitive results, there is room for improvement between weakly- and strong- supervised methods.

\begin{table}[h]
    \centering
    \begin{tabular}{l|x{18}x{18}}
        \textbf{Model} & \textbf{AP$_{50}$} & \textbf{AP$_{75}$}\\\shline
        Base-DA~\cite{dvornik2018importance}& 46.0 & 28.1\\
        Mask-RCNN~\cite{he2017mask}         & 55.2 & 35.3\\\hline
        WISE (Ours)                         & 17.4 & 07.7\\
    \end{tabular}
    \vspace{3mm}
    \caption{\textbf{COCO 2014.} Comparison to fully supervised methods.}
    \label{tab:coco}
\end{table}

\subsection{Experiments on COCO 2014}
For COCO 2014~\cite{lin2014microsoft}, we train on the union of the $80$k train images and the $35$k subset of validation images, and report the results on {\it minival} consisting of $5$k images, following the experimental setup of He~\emph{et al.}~\cite{he2017mask}. It consists of $80$ categories belonging to a wide variety of everyday objects. We obtain ground-truth points by taking the pixel with the largest distance transform for each instance segmentation mask. We use the standard COCO metrics including AP (averaged over IoU thresholds), AP$_{50}$, and AP$_{75}$. Table~\ref{tab:other_baseline} shows that WISE outperforms our baseline ``L-Net + Best Proposal'', which suggests that E-Net generates better proposal masks. The qualitative results in Figure~\ref{fig:qualitative} show that WISE can successfully capture the mask of diverse objects. Table~\ref{tab:coco} shows that while our results are poor compared to fully supervised methods, they establish a first strong baseline for instance segmentation with point-level supervision. 

\subsection{Experiments on KITTI}
KITTI~\cite{Geiger2012CVPR} is a meaningful benchmark for autonomous driving. Using the setup described in \cite{zhang2016instance}, we train our models on the $3,712$ training images where the ground-truth points are the provided bounding box centers. We reported results on the $120$ validation images using the {\it MUCov} and {\it MWCov} metrics, as described in Silberman~\emph{et al.}~\cite{silberman2014instance}. Table~\ref{tab:other_baseline} shows that WISE significantly outperforms the baseline ``L-Net + Best Proposal'', suggesting that relying on the best ``objectness'' score for picking the proposal is not the optimal approach. Furthermore, Table~\ref{tab:kitti} shows that WISE achieves competitive results compared to methods that use full supervision. Figure~\ref{fig:qualitative} shows quality masks being generated for the cars and persons objects on KITTI images by WISE.

\begin{table}[h]
    \centering
    \begin{tabular}{l|x{30}x{30}}
        \textbf{Model} & \textbf{MWCov} & \textbf{MUCov}   \\\shline 
        DepthOrder~\cite{zhang2015monocular}  & 70.9 & 52.2\\ 
        DenseCRF~\cite{zhang2016instance}     & 74.1 & 55.2\\
        AngleFCN+Depth~\cite{uhrig2016pixel}  & 79.7 & 75.8\\ 
        Recurrent+attention~\cite{ren2017end} & 80.0 & 66.9\\\hline
        WISE (Ours)                       & 74.2 & 58.9\\
    \end{tabular}
    \vspace{3mm}
    \caption{\textbf{KITTI.} Comparison to fully supervised methods.}
    \label{tab:kitti}
\end{table}

\subsection{Experiments on CityScapes}
CityScapes~\cite{cordts2016cityscapes} is a popular autonomous driving benchmark for instance segmentation.  It contains $2,975$ high-resolution training images, and $500$ validation images that represent street scenes acquired from an on-board camera. The pixels are labeled into 19 classes, but only 8 classes belong to countable objects (used for instance segmentation): person, rider, car, truck, bus, train, motorcycle, and bicycle. The ground-truth point for each object is the pixel with the largest distance transform within its corresponding ground-truth segmentation mask.

\begin{figure*}[ht]
\begin{center}
\includegraphics[width=0.8\textwidth]{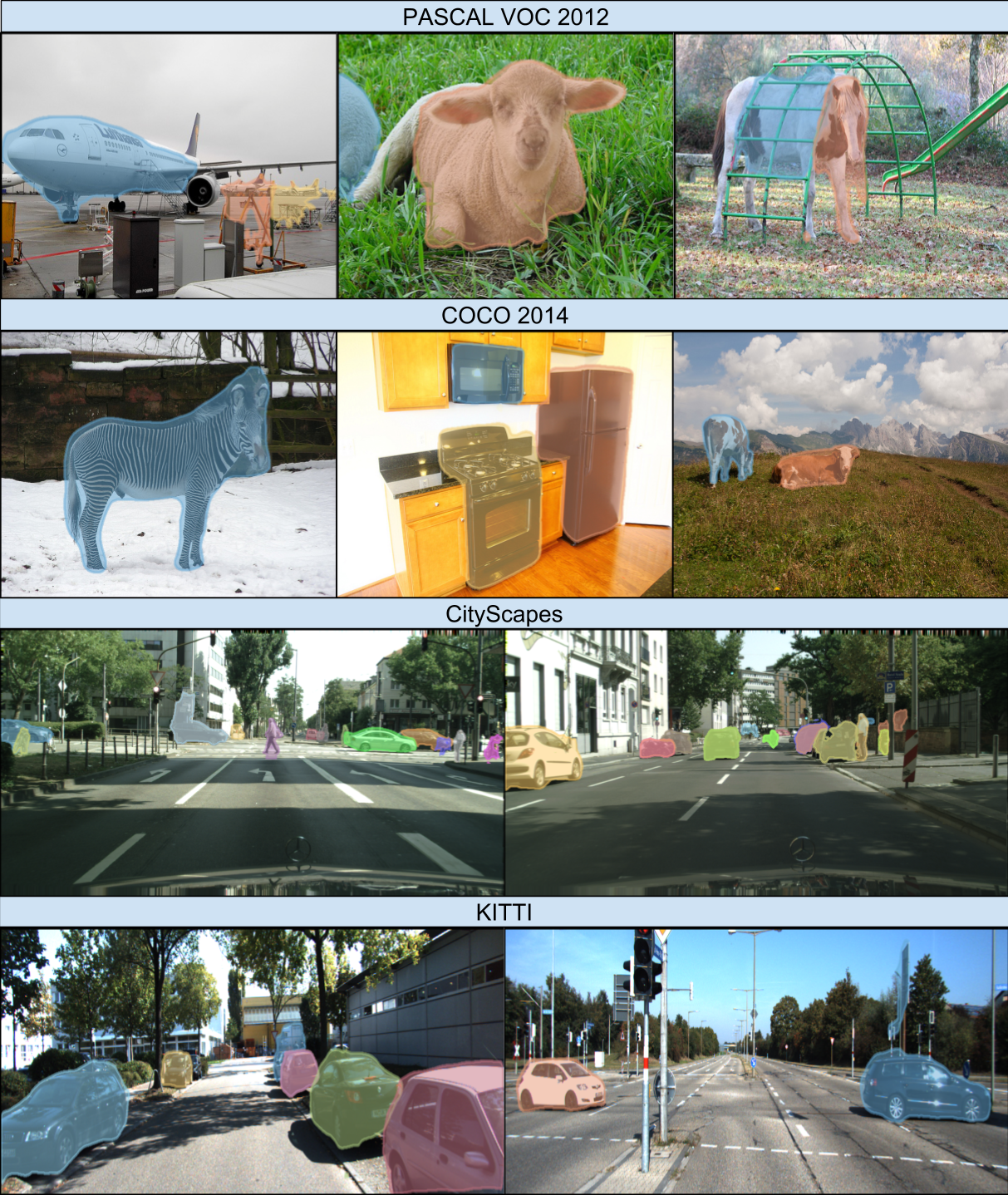}
\end{center}
\caption{\textbf{Qualitative results.} Qualitative results of WISE on the four datasets evaluated.}
\label{fig:qualitative}
\end{figure*}

Table~\ref{tab:other_baseline} shows that WISE sets a new strong baseline for the weakly supervised setting, while achieving better results than the comparable baseline ``L-Net + Best proposal''. Further, Figure~\ref{fig:qualitative} illustrates that our method can obtain good masks for various objects of interest. However, fully supervised methods shown in Table~\ref{tab:cityscapes_full} outperform our weakly supervised method with a large margin, inspiring future research on this problem setup.

In Table~\ref{tab:cityscapes_bbox}, we compare ``GT-points + E-Net'' against the methods proposed by Remez~\emph{et al.}~\cite{remez2018learning} which use bounding box ground-truth labels at test time. Using their evaluation setup, we report the results in Table~\ref{tab:cityscapes_bbox} which shows better results across four categories. This is despite E-Net using weaker labels than Cut \& Paste. According to Bearman~\emph{et al.}~\cite{bearman2016s}, it takes an average of 10.2 seconds to acquire a bounding box, but only 2.4 seconds to get an annotation for a single object instance. 

\begin{table}[h]
    \centering
    \begin{tabular}{l|x{15}}
        \textbf{Method}  & \textbf{AP}\\\shline
        InstanceCut~\cite{kirillov2017instancecut} & 15.8\\
        DWT~\cite{bai2017deep}                     & 19.8\\
        SGN~\cite{liu2017sgn}                      & 29.2\\
        Mask-RCNN~\cite{he2017mask}                & 31.5\\\hline
        WISE (Ours)                                & 07.8\\
    \end{tabular}
    \vspace{3mm}
    \caption{\textbf{CityScapes.} Comparison to fully supervised methods.}
    \label{tab:cityscapes_full}
\end{table}

\begin{table}[h]
    \centering
    \resizebox{\linewidth}{!}{%
    \begin{tabular}{l|x{22}x{22}x{32}x{32}}
        \textbf{Method} & \textbf{Car} & \textbf{Person} & \textbf{T. light} & \textbf{T. sign}\\\shline
        Box~\cite{remez2018learning}             & 62.0 & 49.0 & 76.0 & 76.9\\
        Simple Does it~\cite{khoreva2017simple}  & 68.0 & 53.0 & 60.0 & 51.0\\
        GrabCut~\cite{rother2004grabcut}         & 62.0 & 50.0 & 64.0 & 65.0\\
        Cut \& Paste~\cite{remez2018learning}    & 67.0 & 54.0 & 77.0 & 79.0\\\hline
        Fully Supervised~\cite{remez2018learning}& 80.0 & 61.0 & 79.0 & 81.0\\\hline
        GT-points + E-Net (Ours)                 & 77.6 & 55.4 & 77.8 & 80.1\\
    \end{tabular}
    }
    \vspace{3mm}
    \caption{\textbf{CityScapes.} Methods with bounding boxes at test time.}
    \label{tab:cityscapes_bbox}
\end{table}

\section{Conclusion}
\label{sec:conclusion}
In this paper, we have introduced a weakly supervised instance segmentation network (WISE). It can train by using point-level annotations and by leveraging pseudo-masks from object proposal methods. WISE uses L-Net to first detect the object locations which are  then given as input to E-Net in order to obtain the segmentation masks. E-Net is based on an embedding network that groups pixels in the image-based on their similarity which are then used to select the best matching proposal mask. We have validated our method across a wide variety of datasets. The results show that WISE obtains competitive results against fully supervised methods and outperform weakly-supervised methods with a fixed annotation cost. The results also provide a strong first baseline for instance segmentation with point-level supervision. Although a pretrained proposal method was used in this problem setup, it was not finetuned on any of our datasets. However, an interesting future direction is to address this task with a more challenging setup that requires proposal-free methods.

\clearpage
\clearpage

{\small
\bibliographystyle{ieee}
\bibliography{refs}
}

\end{document}